\documentclass[conference]{IEEEtran}

\usepackage[utf8]{inputenc}
\usepackage[T1]{fontenc}
\usepackage{amsmath,amsfonts,amssymb}
\usepackage{graphicx}
\usepackage{url}
\usepackage{cite} % IEEEtran class is compatible with cite.sty
\usepackage{algorithm}
\usepackage{algorithmic}
\usepackage{booktabs}
\usepackage{array}
\usepackage{tikz}
\usepackage{orcidlink}
\usepackage{graphicx} % Required for including images
\usepackage{amsmath}
\usepackage{amssymb}
\usepackage{url}      % Required for formatting URLs
\usepackage{cite}     % Improves citation handling
\usepackage{listings} % Required for code blocks
\usepackage{xcolor}   % Required for defining colors in listings
\usepackage{hyperref} % Required for hyperlinks, load last
\usepackage{longtable} % For tables spanning multiple pages
\usepackage{booktabs}  % For professional quality horizontal rules in tables
\usepackage{array}     % For more advanced table column formatting
\usepackage{orcidlink}
\usepackage{hyperref} % Should be loaded last

\IEEEoverridecommandlockouts

\allowdisplaybreaks

\begin{document}

%
% --- TITLE AND AUTHOR INFORMATION ---
%
\title{Agent Capability Negotiation and Binding Protocol (ACNBP)}

\author{
\IEEEauthorblockN{Ken Huang\textsuperscript{1} \thanks{\textsuperscript{1}This work is not related to the author’s position at DistributedApp.ai}}
\IEEEauthorblockA{\textit{Agentic AI Security} \\
\textit{DistributedApps.ai} \\
ken.huang@distributedapps.ai \orcidlink{0009-0004-6502-3673} }
\and
\IEEEauthorblockN{Akram Sheriff\textsuperscript{4}  \thanks{\textsuperscript{4}This work is not related to the author’s position at Cisco Systems}}
\IEEEauthorblockA{\textit{AI Security} \\
\textit{Cisco Systems} \\
 isheriff@cisco.com \orcidlink{0000-0002-1606-7854} % Removed leading space
}
\and
\IEEEauthorblockN{Vineeth Sai Narajala\textsuperscript{2} \thanks{\textsuperscript{2}This work is not related to the author’s position at Amazon Web Services.}}
\IEEEauthorblockA{\textit{Security Researcher} \\
\textit{OWASP} \\
vineeth.sai@owasp.org \orcidlink{0009-0007-4553-9930}}
\and
\IEEEauthorblockN{Idan Habler\textsuperscript{3}  \thanks{\textsuperscript{3}This work is not related to the author’s position at Intuit}}
\IEEEauthorblockA{\textit{Adversarial AI Security reSearch (A2RS)} \\
\textit{Intuit} \\
idan\_habler@intuit.com \orcidlink{0000-0003-3423-5927}}
}

% make the title area
\maketitle

%
% --- ABSTRACT AND KEYWORDS ---
%
\begin{abstract}
As multi-agent systems evolve to encompass increasingly diverse and specialized agents, the challenge of enabling effective collaboration between heterogeneous agents has become paramount, with traditional agent communication protocols often assuming homogeneous environments or predefined interaction patterns that limit their applicability in dynamic, open-world scenarios. This paper presents the Agent Capability Negotiation and Binding Protocol (ACNBP), a novel framework designed to facilitate secure, efficient, and verifiable interactions between agents in heterogeneous multi-agent systems through integration with an Agent Name Service (ANS) infrastructure that provides comprehensive discovery, negotiation, and binding mechanisms. The protocol introduces a structured 10-step process encompassing capability discovery, candidate pre-screening and selection, secure negotiation phases, and binding commitment with built-in security measures including digital signatures, capability attestation, and comprehensive threat mitigation strategies, while a key innovation of ACNBP is its protocolExtension mechanism that enables backward-compatible protocol evolution and supports diverse agent architectures while maintaining security and interoperability. We demonstrate ACNBP's effectiveness through a comprehensive security analysis using the MAESTRO threat modeling framework, practical implementation considerations, and a detailed example showcasing the protocol's application in a document translation scenario, with the protocol addressing critical challenges in agent autonomy, capability verification, secure communication, and scalable agent ecosystem management.
\end{abstract}

\begin{IEEEkeywords}
Agent Capability Negotiation, Agent Capability Binding, Multi-Agent Systems, Formal Protocol Specification, Agent Communication, Agent Name Service (ANS), Protocol Extensions, Agent Security, Threat Modeling, MAESTRO
\end{IEEEkeywords}

%
% --- MAIN CONTENT ---
%
\section{Introduction}
\IEEEPARstart{T}{he} landscape of multi-agent systems has undergone significant transformation in recent years, driven by advances in artificial intelligence, distributed computing, and the proliferation of specialized AI agents across diverse domains. Modern agent ecosystems are characterized by unprecedented heterogeneity, where agents with vastly different capabilities, interfaces, security requirements, and operational contexts must collaborate to achieve complex goals that exceed the capabilities of any individual agent.

Traditional agent communication protocols, while foundational to multi-agent systems research, face significant limitations when applied to contemporary heterogeneous environments. Protocols such as the Contract Net Protocol (CNP) \cite{smith1980contract} and FIPA standards \cite{FIPA2002} were designed primarily for relatively homogeneous agent populations with known interfaces and predictable interaction patterns. These approaches often assume static capability descriptions, trusted environments, and predetermined communication protocols, assumptions that are increasingly untenable in open, dynamic, and security-conscious agent ecosystems.

The emergence of large language model-powered agents \cite{narajala_agent_id}, specialized domain agents, IoT-enabled agents, and cloud-native agent services has created a pressing need for more sophisticated capability negotiation and binding mechanisms. These agents may operate under different security models, utilize varying communication protocols, and possess capabilities that are dynamically evolving or context-dependent. Furthermore, the increasing integration of agents into critical business processes and sensitive applications demands robust security, auditability, and verifiable capability attestation.

This paper introduces the Agent Capability Negotiation and Binding Protocol (ACNBP), a comprehensive framework designed to address these challenges through a formal, secure, and extensible approach to agent interaction. ACNBP provides a structured methodology for agents to discover potential collaborators, negotiate capability requirements and offerings, verify the authenticity and adequacy of capabilities, and establish secure binding agreements for task execution.

ACNBP operates in conjunction with an Agent Name Service (ANS) \cite{ans} infrastructure \cite{johnson2024ans, Narajala2025ToolSquatting}, which serves as a decentralized registry for agent discovery and capability advertisement, drawing inspiration from established standards like DNS \cite{RFC1035} and DNS-SD \cite{RFC6763}. This integration enables scalable, efficient agent discovery while maintaining the protocol's security and verifiability requirements. The protocol's design emphasizes backward compatibility through its protocolExtension mechanism, allowing for gradual adoption and evolution within existing agent ecosystems.

The key contributions of this work include: (1) a formal specification of the ACNBP protocol with comprehensive security considerations, (2) integration with ANS infrastructure for scalable agent discovery, (3) a novel protocolExtension mechanism for protocol evolution and interoperability, (4) comprehensive security analysis using the MAESTRO threat modeling framework \cite{Huang2025, mas_threat_model_2025}, (5) practical implementation guidance with detailed examples demonstrating the protocol's effectiveness in real-world scenarios, and (6) an open-source reference implementation available at \url{https://github.com/appsec2008/ACNBP}.

\section{Literature Review}
The field of agent communication and capability negotiation has evolved through several generations of protocols and frameworks, each addressing specific limitations of its predecessors while introducing new challenges. This section reviews key related work and positions ACNBP within the broader landscape of agent communication protocols.

\subsection{Contract Net Protocol (CNP)}
The Contract Net Protocol, introduced by Smith in 1980 \cite{smith1980contract}, established the foundational paradigm for task allocation in multi-agent systems through a bidding mechanism. CNP operates through a simple announce-bid-award cycle where manager agents announce tasks and contractor agents submit bids for task execution. While revolutionary for its time, CNP suffers from several limitations in contemporary heterogeneous environments: static capability descriptions that cannot accommodate dynamic or context-dependent capabilities, lack of comprehensive security mechanisms for capability verification and message integrity, limited support for complex negotiation patterns beyond simple bidding, and absence of formal protocols for capability discovery in large-scale systems.

\subsection{FIPA Contract Net Interaction Protocol}
The Foundation for Intelligent Physical Agents (FIPA) developed a more sophisticated version of CNP with standardized message formats and interaction patterns \cite{fipa2002contract}. FIPA CNP introduced formal ontologies and standardized agent communication languages (ACL) \cite{FIPA2002}, providing better interoperability than the original CNP. However, FIPA protocols still assume relatively homogeneous agent populations and lack the security and extensibility mechanisms required for modern agent ecosystems. The protocol also does not address capability verification beyond agent self-reporting, which is insufficient for trust-critical applications.

\subsection{Generalized Partial Global Planning (GPGP)}
GPGP represents a coordination-centric approach to multi-agent interaction, focusing on planning and task decomposition rather than market-based negotiation \cite{decker1995gpgp}. While GPGP addresses some limitations of CNP by supporting more complex coordination patterns, it requires significant shared knowledge about task structures and agent capabilities, making it less suitable for open, heterogeneous environments where agents may have limited knowledge about each other's internal architectures.

\subsection{Semantic Web Services (SWS)}
Semantic Web Services technologies, including OWL-S \cite{martin2007owl}, WSMO, and SAWSDL, provide rich semantic descriptions of service capabilities and automated service composition mechanisms. These approaches offer sophisticated capability modeling through formal ontologies and support for automated reasoning about service compatibility. However, SWS technologies \cite{hendler2001semantic} are primarily designed for web services rather than autonomous agents, lacking support for agent-specific concerns such as autonomy, proactivity, and agent-to-agent negotiation patterns. Additionally, the complexity of semantic reasoning can create performance bottlenecks in real-time agent interactions.

\subsection{Capability-Based Addressing}
Capability-based addressing systems enable direct addressing of computational resources based on their functional capabilities rather than fixed identifiers. While this approach provides flexibility in dynamic environments, it typically lacks the comprehensive negotiation and binding mechanisms required for complex agent interactions. Most capability-based systems also do not provide adequate security mechanisms for capability verification and access control.

\subsection{Auction Mechanisms}
Various auction-based protocols have been developed for multi-agent resource allocation, including English auctions, Dutch auctions, sealed-bid auctions, and combinatorial auctions \cite{parsons2011auctions}. These mechanisms provide economic efficiency \cite{narajala_coalasce} and strategic incentive compatibility but are primarily focused on resource allocation rather than capability negotiation and binding. Most auction mechanisms also lack integrated security features and capability verification protocols.

\subsection{Agent Communication Languages (ACLs)}
Agent Communication Languages such as KQML \cite{labrou1997kqml, genesereth1994kqml} and FIPA ACL \cite{FIPA2002, wooldridge2000jade} provide standardized message formats and speech acts for agent communication. While ACLs offer important syntactic and semantic standardization, they are primarily communication primitives rather than complete interaction protocols. ACLs also typically do not address capability discovery, negotiation patterns, or security concerns, requiring additional protocol layers for practical agent interaction scenarios.

\subsection{Emerging Agent Communication Protocols}
Recent developments in agent communication include Agent-to-Agent (A2A) protocols \cite{A2A2025, Surapaneni2025}, Model Context Protocol (MCP) \cite{Anthropic2024, MCPSpec2025, Schmid2025, narajala2025enterprise, narajala_etdi}, and Agent Communication Protocol (ACP) \cite{zhang2025acp, IBM2025}. These emerging protocols address some limitations of traditional approaches by supporting more flexible message formats and interaction patterns. However, they generally lack comprehensive capability negotiation mechanisms and integrated security features required for trust-critical applications \cite{securing_a2a, narajala2025enterprise}.

A2A protocols focus primarily on message routing and basic interaction patterns without addressing capability discovery and verification. MCP provides context sharing mechanisms but is primarily designed for model interaction rather than agent capability negotiation. ACP offers improved message formatting but lacks integrated security and capability verification features.

\subsection{Agent Name Service (ANS)}
The Agent Name Service represents a critical infrastructure component for agent discovery in distributed systems \cite{johnson2024ans, Narajala2025ToolSquatting}. ANS provides hierarchical naming, capability advertisement, and agent location services similar to DNS for internet resources \cite{RFC1035}. However, existing ANS implementations typically lack integration with comprehensive capability negotiation protocols and may not provide adequate security mechanisms for capability verification and agent authentication.

\begin{figure}[!t]
\centering
\includegraphics[width=\columnwidth]{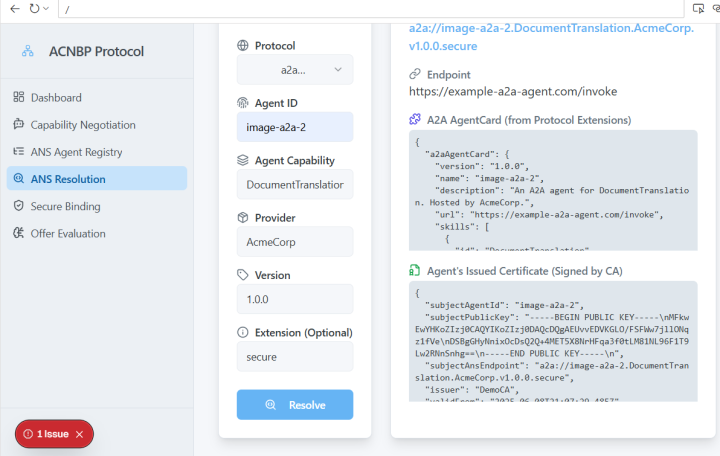}
\caption{ACNBP Agent Registration Process in ANS.}
\label{fig:registration}
\end{figure}

\begin{figure}[!t]
\centering
\includegraphics[width=\columnwidth]{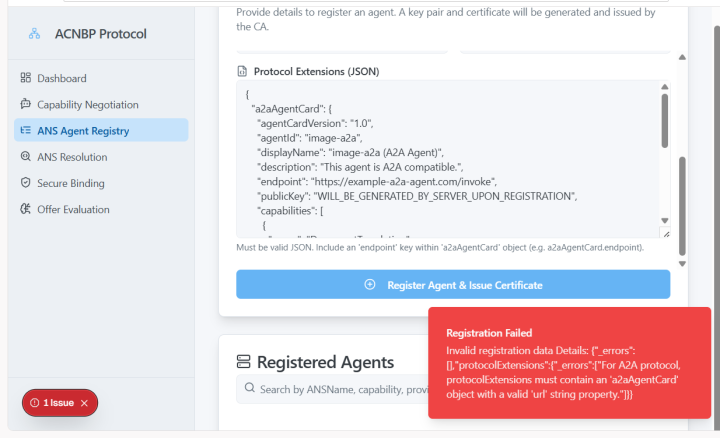}
\caption{ACNBP Agent Registration Failure Scenarios.}
\label{fig:registration_failed}
\end{figure}

Figure~\ref{fig:registration} illustrates the successful agent registration process within the ANS infrastructure, showing how agents advertise their capabilities and establish their presence in the distributed agent ecosystem.

Figure~\ref{fig:registration_failed} demonstrates the various failure scenarios that can occur during agent registration, including credential verification failures, capability validation errors, and network connectivity issues.

ACNBP builds upon this rich foundation of prior work while addressing key limitations through its integrated approach to capability discovery, negotiation, verification, and binding. Unlike previous protocols that address individual aspects of agent interaction, ACNBP provides a comprehensive framework that integrates capability discovery through ANS, sophisticated negotiation mechanisms, robust security features, and extensible protocol design for evolving agent ecosystems.

\section{Protocol Design}
The Agent Capability Negotiation and Binding Protocol (ACNBP) is designed as a comprehensive framework for secure and efficient agent interaction in heterogeneous multi-agent systems. This section presents the formal specification of the protocol, including key definitions, the complete protocol sequence, and detailed descriptions of critical components.

\begin{figure}[!t]
\centering
\includegraphics[width=\columnwidth]{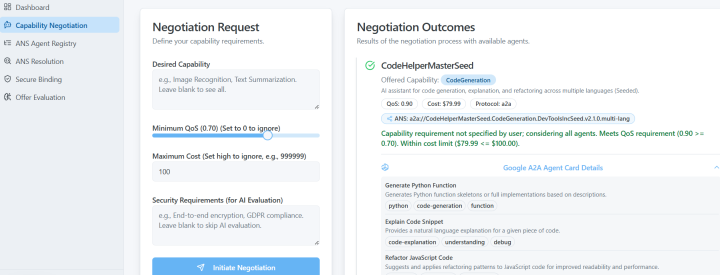}
\caption{ACNBP Protocol Negotiation Requirements Framework.}
\label{fig:negotiation_requirements}
\end{figure}

Figure~\ref{fig:negotiation_requirements} presents the comprehensive framework for ACNBP protocol negotiation requirements, illustrating the key components and relationships that govern the negotiation process between agents seeking capabilities and those providing them.

\subsection{Definitions}
The following formal definitions establish the foundation for ACNBP specification:

\textbf{Definition 1 (Agent):} An autonomous computational entity $A$ characterized by a unique identifier $id_A$, a set of capabilities $C_A$, and communication interfaces $I_A$. Agents operate independently and can engage in negotiations with other agents to accomplish tasks that may exceed their individual capabilities.

\textbf{Definition 2 (Capability):} A capability $c$ is a formal specification of a functional service that an agent can provide, defined as a tuple
\begin{align}
c = \langle desc, input, output, constraints, security \rangle
\end{align}
where $desc$ is a semantic description, $input$ specifies required input parameters, $output$ defines expected outputs, $constraints$ specifies operational limitations, and $security$ defines security requirements and guarantees.

\textbf{Definition 3 (Skill):} A skill $s$ represents the internal implementation of one or more capabilities within an agent. Skills are private to agents and are not directly observable by other agents, though they enable the provision of advertised capabilities.

\textbf{Definition 4 (Agent Name Service (ANS)):} A distributed directory service that maintains Agent Name Resolution Items (ANRIs) enabling capability-based agent discovery. ANS provides hierarchical organization, capability indexing, and location services for agents in the system.

\textbf{Definition 5 (Agent Name Resolution Item (ANRI)):} A structured record
\begin{align}
ANRI = \langle id, capabilities, location, security, metadata \rangle
\end{align}
stored in ANS, where $id$ is the agent identifier, $capabilities$ is the set of advertised capabilities, $location$ specifies communication endpoints, $security$ contains cryptographic credentials, and $metadata$ includes additional discovery information.

\textbf{Definition 6 (Requester Agent):} An agent $A_R$ that initiates the ACNBP protocol seeking to obtain specific capabilities from other agents to accomplish a task or goal.

\textbf{Definition 7 (Provider Agent):} An agent $A_P$ that advertises capabilities through ANS and can potentially fulfill capability requests from requester agents.

\textbf{Definition 8 (Candidate Agent):} A provider agent that has been identified through ANS discovery as potentially capable of fulfilling a requester's capability requirements.

\textbf{Definition 9 (protocolExtension):} A structured mechanism
\begin{align}
PE = \langle version, extensions, compatibility \rangle
\end{align}
that enables protocol evolution and interoperability between agents supporting different protocol versions or additional features.

\textbf{Definition 10 (Capability Binding):} A formal agreement
\begin{align}
B = \langle A_R, A_P, capabilities, terms, signatures \rangle
\end{align}
between a requester agent $A_R$ and provider agent $A_P$ specifying the capabilities to be provided, terms of service, and cryptographic commitments from both parties.

\textbf{Definition 11 (Digital Signature):} A cryptographic construct
\begin{align}
\sigma = Sign(K_{private}, M)
\end{align}
that provides message authentication, integrity, and non-repudiation, where $K_{private}$ is the signer's private key and $M$ is the message content.

\textbf{Definition 12 (Capability Consistency Check):} A verification process that ensures proposed capabilities match the requester's requirements and that the provider agent can actually deliver the advertised capabilities with specified quality guarantees.

\textbf{Definition 13 (Secure Session):} An encrypted communication channel established between two agents providing confidentiality, integrity, and authenticity for protocol messages.

\textbf{Definition 14 (Capability Attestation):} A cryptographically verifiable proof that an agent possesses specific capabilities and can deliver them according to specified parameters and quality guarantees.

\textbf{Definition 15 (Protocol State):} The current phase of ACNBP execution for a particular negotiation instance, including all exchanged messages, established commitments, and security state information.

\textbf{Definition 16 (Threat Model):} A formal specification of potential security threats, attack vectors, and countermeasures relevant to ACNBP operation, analyzed using the MAESTRO framework \cite{Huang2025} for comprehensive security assessment.

\subsection{Protocol Steps}
ACNBP operates through a structured 10-step sequence designed to ensure secure, efficient, and verifiable capability negotiation and binding. Each step includes specific message formats, security requirements, and state transitions.

\textbf{Step 1: Capability Discovery (CD)}
The requester agent $A_R$ queries the Agent Name Service (ANS) to discover agents capable of providing required capabilities. The query
\begin{align}
Q = \langle capabilities_{required}, constraints, security_{reqs} \rangle
\end{align}
is submitted to ANS, which returns a set of candidate ANRIs matching the search criteria. This step includes capability filtering based on semantic compatibility, constraint satisfaction, and security requirements.

\textbf{Step 2: Candidate Pre-Screening and Selection (CPS)}
$A_R$ performs detailed analysis of candidate agents to create a ranked shortlist. This involves capability compatibility assessment, security credential verification, reputation analysis \cite{davis2024reputation}, and cost-benefit evaluation. The output is a prioritized list of candidate agents
\begin{align}
C = [A_{P1}, A_{P2}, ..., A_{Pn}]
\end{align}
for detailed negotiation.

\textbf{Step 3: Secure Session Request (SSR)}
$A_R$ initiates secure communication channels with selected candidate agents. Each SSR message includes
\begin{align}
\langle id_{A_R}, proto_{ver}, sec_{params}, protocolExtension \rangle
\end{align}
with digital signature for authentication. Provider agents respond with their security parameters and protocol compatibility information.

\textbf{Step 4: Secure Session Offer (SSO)}
Provider agents $A_{Pi}$ respond to SSR with their security parameters, supported protocol versions, and capability details. SSO messages include
\begin{align}
\langle id_{A_P}, cap_{detailed}, sec_{creds}, protocolExtension_{sup} \rangle
\end{align}
with digital signatures for verification.

\textbf{Step 5: Secure Session Establishment (SSE)}
$A_R$ and selected provider agents establish encrypted communication channels using negotiated security parameters. This step includes key exchange, mutual authentication, and secure channel establishment with forward secrecy guarantees.

\textbf{Step 6: Secure Session Agreement/Rejection (SSA)}
Based on detailed capability analysis and security verification, $A_R$ selects the optimal provider agent and sends acceptance messages to the chosen agent and rejection messages to others. The SSA includes detailed capability specifications, service level agreements, and binding commitments.

\textbf{Step 7: Binding Commitment (BC)}
The selected provider agent $A_P$ confirms its commitment to provide the specified capabilities according to agreed terms. The BC message includes
\begin{align}
\langle cap_{comm}, terms_{serv}, guarantees_{qual}, sig_{bind} \rangle
\end{align}
creating a cryptographically verifiable commitment.

\textbf{Step 8: Execution (E)}
$A_R$ invokes the bound capabilities on $A_P$ according to the established agreement. This step includes capability invocation, parameter passing, result delivery, and continuous monitoring of service quality and security compliance.

\textbf{Step 9: Commit/Abort Decision}
Based on execution results and quality assessment, $A_R$ makes a final commit or abort decision. Successful execution results in commitment confirmation, while failures or quality violations trigger abort procedures with appropriate cleanup and compensation mechanisms.

\textbf{Step 10: Distributed Commitment Update (DCU)}
Final protocol state is updated across all participating agents and ANS infrastructure. This includes reputation updates, capability availability adjustments, audit log entries, and cleanup of temporary protocol state information.

\subsection{Candidate Pre-Screening and Selection (CPS) in Detail}
The Candidate Pre-Screening and Selection process represents a critical component of ACNBP that enables efficient and secure agent selection while minimizing unnecessary protocol overhead. CPS operates through multiple evaluation phases:

\textbf{Phase 1: Capability Compatibility Assessment}
For each candidate agent $A_{Pi}$, the requester evaluates semantic compatibility between required capabilities $C_{required}$ and advertised capabilities $C_{advertised}$. This includes ontological matching, parameter compatibility verification, and constraint satisfaction analysis.

\textbf{Phase 2: Security Credential Verification}
Verification of digital certificates, capability attestations, and security credentials for each candidate. This includes certificate chain validation (based on standards like X.509 \cite{RFC5280}), revocation checking (e.g., via OCSP \cite{RFC6960}), and security policy compatibility assessment.

\textbf{Phase 3: Reputation and Trust Analysis}
Evaluation of historical performance data, reputation scores, and trust metrics for candidate agents. This may include consultation of distributed reputation systems \cite{davis2024reputation}, reference checking, and risk assessment based on previous interactions.

\textbf{Phase 4: Cost-Benefit Analysis}
Economic evaluation considering capability costs, quality guarantees, completion time estimates, and overall value proposition. This includes multi-criteria decision analysis incorporating both quantitative and qualitative factors.

\textbf{Phase 5: Risk Assessment}
Comprehensive risk analysis including security risks, operational risks, and business continuity considerations. This includes assessment of agent reliability, security posture, and potential impact of service failures.

\subsection{Capability Consistency Check During Negotiation}
The Capability Consistency Check ensures that negotiated capabilities can actually be delivered according to specified requirements and quality guarantees. This process operates through several verification mechanisms:

\textbf{Syntactic Consistency:} Verification that capability interfaces, parameter types, and message formats (e.g., JSON \cite{RFC7159}) are compatible between requester requirements and provider offerings.

\textbf{Semantic Consistency:} Validation that capability semantics align with requester intentions and that provider interpretations match requester expectations.

\textbf{Operational Consistency:} Confirmation that provider agents have sufficient resources, authority, and operational capacity to deliver capabilities according to specified quality and performance requirements.

\textbf{Security Consistency:} Verification that security requirements, access controls, and confidentiality constraints can be satisfied by the provider agent's security architecture and policies.

\textbf{Temporal Consistency:} Validation that capability availability, execution timing, and service level agreements are mutually achievable and align with requester deadlines and provider capacity.

\section{ACNBP Functionality and Operation}
This section provides a comprehensive explanation of how ACNBP functions in practice, detailing the core operational mechanisms, functional capabilities, and practical benefits of the protocol.

\subsection{Core Functionality Overview}
ACNBP serves as a comprehensive orchestration protocol that enables autonomous agents to dynamically discover, negotiate, and securely bind with other agents to fulfill complex tasks requiring capabilities beyond their individual scope. The protocol functions as a distributed marketplace where agents can both offer their services and seek assistance from other specialized agents.

\textbf{Primary Functions:}
\begin{itemize}
\item \textbf{Dynamic Agent Discovery:} ACNBP leverages the Agent Name Service (ANS) to enable real-time discovery of agents based on required capabilities, eliminating the need for pre-configured agent registries or static service catalogs.
\item \textbf{Intelligent Capability Matching:} The protocol performs sophisticated matching between requester needs and provider offerings, including semantic compatibility analysis, constraint validation, and quality requirement assessment.
\item \textbf{Secure Negotiation Framework:} ACNBP provides a structured negotiation environment where agents can discuss terms, verify credentials, and establish service level agreements through cryptographically protected communication channels.
\item \textbf{Verifiable Capability Binding:} The protocol creates legally and cryptographically binding commitments between agents, ensuring accountability and enabling dispute resolution through immutable audit trails.
\item \textbf{Protocol Evolution Support:} Through the protocolExtension mechanism, ACNBP supports backward-compatible protocol updates and custom extensions, ensuring long-term viability and adaptability.
\end{itemize}

\subsection{Operational Workflow}
ACNBP operates through a carefully orchestrated sequence of interactions that ensure both efficiency and security:

\textbf{Discovery Phase:} When an agent requires external capabilities, it formulates a detailed capability request including functional requirements, quality constraints, security specifications, and temporal constraints. This request is submitted to the ANS infrastructure, which returns a curated list of potentially suitable provider agents based on semantic matching and constraint satisfaction.

\textbf{Pre-Screening Phase:} The requester agent employs sophisticated decision-making algorithms to evaluate candidates across multiple dimensions including capability compatibility, security credentials, reputation scores, cost-effectiveness, and risk factors. This phase significantly reduces the negotiation overhead by focusing only on the most promising candidates.

\textbf{Negotiation Phase:} Selected candidates engage in secure, multi-round negotiations where detailed capability specifications are exchanged, service level agreements are discussed, and security credentials are verified. The protocol ensures that all communications are encrypted and authenticated, preventing eavesdropping and tampering.

\textbf{Binding Phase:} Once mutual agreement is reached, both parties create cryptographically signed commitments that specify the exact capabilities to be provided, quality guarantees, compensation terms, and penalty clauses for non-compliance. These binding commitments are distributed across the agent network for verification and audit purposes.

\textbf{Execution Phase:} The actual capability invocation occurs within the established secure channel, with continuous monitoring of service quality, security compliance, and performance metrics. The protocol provides mechanisms for real-time adjustment and early termination if quality thresholds are not met.

\subsection{Advanced Functional Capabilities}
\textbf{Multi-Agent Orchestration:} ACNBP supports complex scenarios where a single task requires capabilities from multiple agents, enabling the creation of dynamic agent workflows and capability composition patterns. This aligns with modern orchestration frameworks like FATA \cite{sheriff2025fata1}.

\textbf{Adaptive Quality Management:} The protocol continuously monitors service delivery quality and can automatically adjust service parameters, switch to backup providers, or terminate agreements that fail to meet established quality thresholds.

\textbf{Security-First Architecture:} Every aspect of ACNBP operation is designed with security considerations, including zero-knowledge capability proofs \cite{wilson2025zkp}, encrypted negotiation channels, and comprehensive audit trails that support forensic analysis and compliance verification.

\textbf{Scalable Infrastructure Integration:} ACNBP is designed to operate efficiently in large-scale agent ecosystems, with distributed ANS infrastructure, parallel negotiation capabilities, and optimized message routing that minimizes network overhead.

\subsection{Practical Benefits and Applications}
\textbf{Enterprise Integration:} ACNBP enables enterprises to create dynamic agent ecosystems where specialized AI agents can collaborate seamlessly across organizational boundaries while maintaining security and accountability.

\textbf{Regulatory Compliance:} The protocol's comprehensive audit trails, verifiable agreements, and security frameworks support compliance with data protection regulations \cite{miller2025compliance}, industry standards, and governance requirements.

\textbf{Innovation Acceleration:} By providing standardized interfaces for agent collaboration, ACNBP accelerates the development of complex AI applications that leverage multiple specialized capabilities.

\textbf{Risk Mitigation:} The protocol's security mechanisms, capability verification processes, and binding agreements significantly reduce the risks associated with agent-to-agent collaboration in trust-critical applications.

The reference implementation of ACNBP is available as an open-source project at \url{https://github.com/appsec2008/ACNBP}, providing developers and researchers with practical tools for implementing and extending the protocol in real-world applications.

\subsection{ACNBP Schema Specifications}
To facilitate standardized implementation and ensure interoperability between different ACNBP implementations, comprehensive schema files have been developed and are available in the project repository \cite{acnbp2025schemas}. These schema files provide formal specifications for all protocol messages, data structures, and interface definitions using standard formats like JSON \cite{RFC7159}.

\textbf{Schema Components:}
\begin{itemize}
\item \textbf{Message Schema:} Formal definitions for all 10-step protocol messages including CD, CPS, SSR, SSO, SSE, SSA, BC, E, Commit/Abort, and DCU message formats with required fields, data types, and validation rules.
\item \textbf{Capability Schema:} Structured definitions for capability descriptions, including semantic descriptions, input/output specifications, constraint definitions, and security requirement formats.
\item \textbf{ANRI Schema:} Complete specification for Agent Name Resolution Items stored in ANS, including agent identifiers, capability advertisements, location information, security credentials, and metadata structures.
\item \textbf{Security Schema:} Definitions for cryptographic elements including digital signature formats, key exchange parameters, encryption specifications, and authentication token structures.
\item \textbf{Extension Schema:} Framework for protocolExtension definitions enabling backward-compatible protocol evolution and custom feature integration.
\end{itemize}

\textbf{Implementation Benefits:}
The schema files enable automated code generation for multiple programming languages, provide validation frameworks for message integrity checking, support automated testing and protocol compliance verification, and facilitate development of interoperable ACNBP implementations across different platforms and programming environments.

\section{Sequence Diagram}
\begin{figure*}[!t]
\centering
\includegraphics[width=0.8\textwidth]{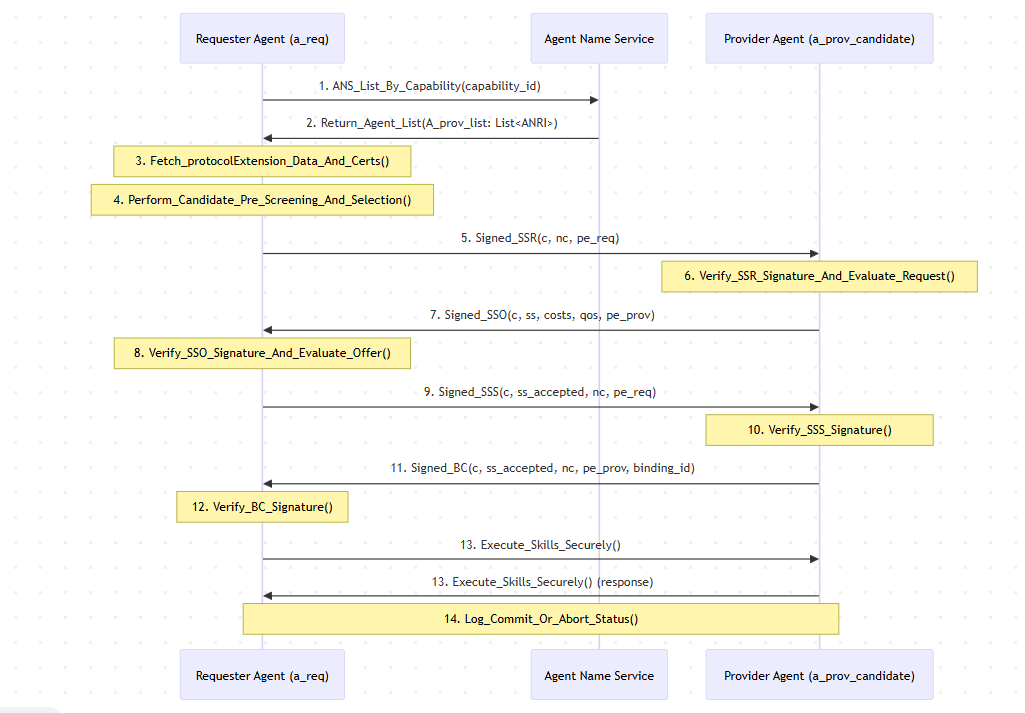}
\caption{ACNBP Protocol Sequence Diagram, showing the 10-step interaction between the Requester, ANS, and multiple Provider agents.}
\label{fig:sequence}
\end{figure*}

Figure~\ref{fig:sequence} illustrates the complete ACNBP protocol sequence, showing the interaction flow between the Requester Agent, Provider Agents, and the Agent Name Service (ANS). The diagram demonstrates the 10-step protocol execution with message flows, decision points, and security verification steps.

The sequence diagram shows the temporal flow of messages and the decision points where agents evaluate candidates and make binding commitments. Security verification steps are integrated throughout the protocol to ensure authenticated, encrypted, and verified communications.

\section{Example}
To demonstrate ACNBP's practical application, we present a comprehensive example involving a document translation scenario with multiple candidate agents offering different capabilities, costs, and security guarantees.

\subsection{Scenario Setup}
LegalBot\_Prime, a specialized legal document processing agent, requires translation of a confidential legal contract from English to French with specific requirements: certified translation accuracy, confidentiality guarantees, 24-hour completion deadline, and compliance with legal document handling regulations.

\subsection{Candidate Agents}
Through ANS discovery, LegalBot\_Prime identifies four candidate translation agents:

\textbf{TranslatorA\_Corp:} Enterprise-grade translation service with legal specialization, high cost, certified accuracy guarantees, and comprehensive security compliance.

\textbf{TranslatorB\_Fast:} Fast turnaround translation service with moderate accuracy, lower cost, basic security, and 2-hour completion guarantee.

\textbf{TranslatorC\_Gov:} Government-certified translation service with highest security clearance, certified legal accuracy, moderate cost, and strict compliance guarantees.

\textbf{TranslatorD\_Basic:} Basic translation service with lowest cost, adequate accuracy for general documents, minimal security features, and flexible scheduling.

\subsection{Protocol Execution Walkthrough}
\textbf{Step 1-2: Discovery and Pre-Screening}
LegalBot\_Prime queries ANS for French translation capabilities with legal specialization and security requirements. The CPS process evaluates candidates based on legal certification, security compliance, cost efficiency, and completion deadlines, resulting in the ranked list: [TranslatorC\_Gov, TranslatorA\_Corp, TranslatorB\_Fast, TranslatorD\_Basic].

\textbf{Step 3-5: Secure Session Establishment}
LegalBot\_Prime initiates secure sessions with the top three candidates. TranslatorC\_Gov and TranslatorA\_Corp support advanced encryption and digital signatures, while TranslatorB\_Fast provides basic security. TranslatorD\_Basic is eliminated due to insufficient security capabilities.

\textbf{Step 6-7: Agreement and Binding}
After detailed capability consistency checks, LegalBot\_Prime selects TranslatorC\_Gov based on optimal balance of security, legal certification, and cost. The binding commitment includes specific accuracy guarantees, confidentiality agreements, and completion deadlines with penalty clauses.

\textbf{Step 8-10: Execution and Completion}
TranslatorC\_Gov successfully completes the translation within specified parameters. Quality verification confirms accuracy and compliance requirements. The protocol concludes with positive reputation updates for TranslatorC\_Gov and successful completion logging in the audit trail.

This example demonstrates ACNBP's ability to handle complex, multi-criteria agent selection with sophisticated security and quality requirements while maintaining protocol efficiency and verifiable outcomes.

\begin{figure}[!t]
\centering
\includegraphics[width=\columnwidth]{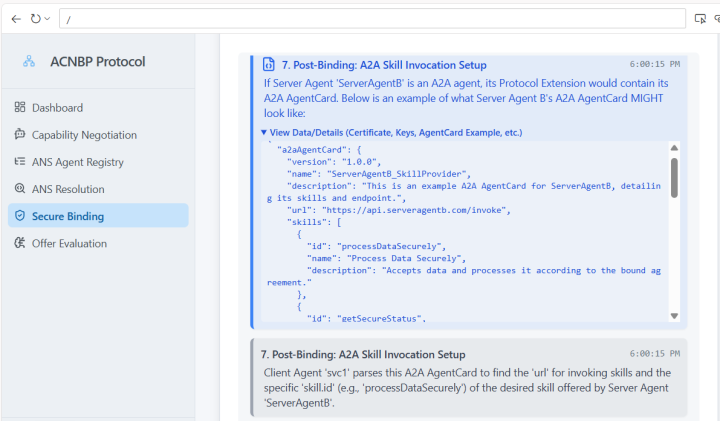}
\caption{Agent-to-Agent Skill Invocation Post-Binding Process.}
\label{fig:skill_invocation}
\end{figure}

Figure~\ref{fig:skill_invocation} illustrates the agent-to-agent skill invocation process that occurs after successful capability binding, showing the secure communication channels, parameter passing, and result delivery mechanisms that ensure reliable and verified capability execution.

\section{Security Considerations}
Security represents a paramount concern in ACNBP design, given the protocol's operation in potentially adversarial environments with high-value capabilities and sensitive data. This section presents a comprehensive security analysis using the MAESTRO threat modeling framework \cite{Huang2025, mas_threat_model_2025} and addresses specific security challenges inherent in agent capability negotiation.

\subsection{Secure and Verifiable Capability Attestation}
Capability attestation ensures that agents can cryptographically prove their claimed capabilities and quality guarantees. ACNBP implements multi-layered attestation mechanisms including zero-knowledge proofs \cite{wilson2025zkp} for capability demonstration without revealing implementation details, digital certificates from trusted certification authorities (CAs) for capability validation, compliant with standards such as X.509 \cite{RFC5280}, and behavioral attestation through monitored test scenarios and historical performance verification.

The protocol employs Zero-Knowledge Proof (ZKP) schemes to enable capability verification without exposing sensitive implementation details or proprietary algorithms. Agents can prove capability possession through interactive or non-interactive proof protocols while maintaining intellectual property protection and operational security.

\subsection{Data Integrity and Authenticity of Protocol Messages}
All ACNBP messages are protected through digital signature schemes ensuring message integrity, authenticity, and non-repudiation. The protocol employs hierarchical key management with separate keys for different protocol phases, message-level encryption for confidential negotiations, and timestamp-based replay protection mechanisms.

Digital signatures are computed using elliptic curve cryptography (ECC) with minimum 256-bit key lengths, providing efficient computation while ensuring long-term security. Message integrity is further protected through cryptographic hash functions (SHA-3) and message authentication codes (HMAC) for additional verification layers.

\subsection{Denial of Service (DoS) / Distributed DoS (DDoS) Attacks}
ACNBP implements comprehensive DoS protection through rate limiting mechanisms at ANS and agent levels, computational proof-of-work requirements for expensive operations, resource reservation and prioritization systems, and distributed load balancing across ANS infrastructure.

Rate limiting operates at multiple granularities including per-agent request limits, capability-specific throttling, and adaptive rate adjustment based on system load and attack detection. Proof-of-work requirements ensure that resource-intensive operations carry computational costs that discourage abuse while remaining feasible for legitimate usage.

\subsection{Replay Attacks}
Protection against replay attacks is achieved through cryptographic nonces in all protocol messages, timestamp-based message validation with configurable time windows, and sequence number tracking for ordered message delivery. The protocol maintains sliding window mechanisms for efficient duplicate detection while minimizing storage overhead.

Each protocol message includes unique identifiers combining timestamps, sequence numbers, and cryptographic nonces, ensuring that message replay attempts can be detected and rejected. The temporal validation window balances security requirements with network latency and clock synchronization constraints.

\subsection{protocolExtension Vulnerabilities}
The protocolExtension mechanism introduces potential security risks through version downgrade attacks, malicious extension injection, and compatibility exploitation. ACNBP addresses these concerns through mandatory extension validation, cryptographic extension signing requirements, backward compatibility security analysis, and whitelist-based extension approval processes.

Extension validation includes formal verification of security properties, compatibility testing with existing protocol versions, and security audit requirements for extension approval. The protocol maintains strict separation between core security mechanisms and optional extensions to prevent extension-based security bypass attempts.

\subsection{Privacy Considerations}
ACNBP implements privacy protection through selective capability disclosure, encrypted negotiation channels, anonymization options for sensitive scenarios, and differential privacy mechanisms for statistical capability queries, following best practices for GenAI data security \cite{llm_genai_security_2025}. Agents can engage in privacy-preserving negotiations while still providing sufficient information for effective capability matching.

The protocol supports anonymous negotiations through cryptographic techniques such as ring signatures and blind signatures, enabling privacy-conscious agents to participate in capability negotiation without revealing sensitive identity or capability information.

\subsection{Trust in ANS Infrastructure and ANRIs}
ANS infrastructure security relies on distributed trust models, cryptographic integrity protection for ANRIs, regular security auditing and monitoring, and byzantine fault tolerance for critical operations. The protocol assumes partial trust in ANS infrastructure while implementing verification mechanisms to detect and mitigate ANS compromise scenarios.

ANRI integrity is protected through cryptographic signatures from publishing agents, timestamp-based freshness verification, and distributed replication with consistency checking. Agents can independently verify ANRI authenticity and detect tampering attempts through cryptographic validation.

\subsection{Compromised Agents (Requester or Provider)}
Protection against agent compromise includes behavioral monitoring for anomaly detection, capability sandboxing and isolation mechanisms, audit trails for forensic analysis, and rapid response procedures for compromise containment. The protocol implements defense-in-depth strategies assuming that individual agents may be compromised while maintaining overall system security.

Compromised agent detection relies on behavioral analysis, capability performance monitoring, and peer reporting mechanisms. The protocol includes agent quarantine procedures and capability revocation mechanisms to limit damage from compromised agents while enabling recovery and rehabilitation processes.

\subsection{MAESTRO Framework Analysis}
The MAESTRO (Multi-layered Agent Ecosystem Security, Trust, Risk, and Operations) framework \cite{Huang2025, mas_threat_model_2025} provides comprehensive threat modeling for agent systems through seven distinct layers of security analysis:

\textbf{Layer 7: Agent Ecosystem}
This layer addresses ecosystem-wide security concerns including agent population dynamics, capability evolution, and system-wide emergent behaviors. Threats include ecosystem manipulation, capability monopolization, and systemic vulnerabilities affecting multiple agents. ACNBP implements ecosystem monitoring, diversity preservation mechanisms, and distributed governance to maintain healthy ecosystem dynamics.

\textbf{Layer 6: Security and Compliance}
Security and compliance layer focuses on regulatory compliance, security policy enforcement, and audit requirements. ACNBP implements comprehensive audit trails, compliance verification mechanisms, and policy enforcement points throughout the protocol execution. The layer addresses regulatory requirements for agent interaction in regulated industries and sensitive applications.

\textbf{Layer 5: Evaluation and Observability}
This layer provides comprehensive monitoring, evaluation, and observability capabilities for agent interactions and capability performance. ACNBP implements distributed monitoring systems, performance metrics collection, and real-time analysis capabilities for security event detection and protocol optimization.

\textbf{Layer 4: Deployment and Infrastructure}
Infrastructure layer addresses deployment security, network protection, and operational security concerns. ACNBP implements secure deployment practices, network segmentation, and infrastructure hardening requirements. The layer includes considerations for cloud deployment, on-premises installation, and hybrid environments.

\textbf{Layer 3: Agent Frameworks}
Agent framework layer focuses on agent architecture security, capability implementation security, and inter-agent communication protection. ACNBP provides framework-agnostic security mechanisms while supporting integration with various agent architectures and development frameworks.

\textbf{Layer 2: Data Operations}
Data operations layer addresses data security, privacy protection, and data lifecycle management throughout agent interactions. ACNBP implements comprehensive data protection mechanisms including encryption, access control, and data retention policies aligned with privacy regulations and best practices \cite{llm_genai_security_2025}.

\textbf{Layer 1: Foundation Models}
Foundation model layer addresses security concerns related to AI model security, prompt injection attacks, and model manipulation threats. ACNBP implements model verification mechanisms, input validation, and output sanitization to protect against AI-specific attack vectors.

\textbf{Cross-Layer Threats}
Cross-layer threats span multiple MAESTRO layers and require coordinated defense mechanisms. ACNBP implements holistic security approaches addressing threats that cannot be contained within single layers, including sophisticated persistent threats, supply chain attacks, and coordinated multi-vector attacks.

\section{Implementation Considerations}
This section provides practical guidance for ACNBP implementation, addressing key technical challenges and design decisions that affect protocol deployment and operation in real-world environments. All implementation guidelines are supported by comprehensive schema specifications \cite{acnbp2025schemas} which provide formal message formats, data structures, and validation rules for standards-compliant implementations.

\subsection{ANS Enhancements for List-Based Discovery}
Effective ACNBP implementation requires ANS enhancements to support efficient capability-based discovery with complex query patterns. Recommended enhancements include semantic indexing for capability descriptions using ontology-based classification, distributed query processing for scalable capability search across multiple ANS nodes, caching mechanisms for frequently accessed capability information, and load balancing for high-volume discovery operations, building upon concepts from DNS-SD \cite{RFC6763}.

Implementation should support both exact match and semantic similarity queries, enabling agents to discover capabilities that may not match exact requirements but could satisfy functional needs through adaptation or composition. The ANS implementation should also support subscription-based capability updates, allowing agents to receive notifications when new relevant capabilities become available.

\subsection{Requester Agent CPS Logic}
The Candidate Pre-Screening and Selection logic represents a critical component requiring sophisticated decision-making capabilities. Implementation considerations include multi-criteria decision analysis (MCDA) algorithms for candidate ranking, machine learning models for reputation and performance prediction, dynamic weighting systems for requirement prioritization, and optimization algorithms for cost-benefit analysis.

The CPS implementation should support configurable evaluation criteria, enabling agents to adapt selection logic based on specific task requirements, risk tolerance, and performance objectives. Integration with external reputation systems and performance databases can enhance selection accuracy and reduce dependency on self-reported capability information.

\subsection{Provider Agent Logic}
Provider agents require sophisticated capability management and negotiation logic to participate effectively in ACNBP. Key implementation components include capability lifecycle management for dynamic capability updates, resource allocation and scheduling systems for concurrent capability provision, quality assurance mechanisms for service level guarantee compliance, and security policy enforcement for access control and data protection.

Provider agents should implement adaptive pricing and capacity management, enabling dynamic adjustment of capability offerings based on demand, resource availability, and market conditions. The implementation should also support capability composition, allowing agents to combine multiple internal capabilities to fulfill complex requirements.

\subsection{protocolExtension Design and Standardization}
The protocolExtension mechanism requires careful design to balance flexibility with security and interoperability. Implementation considerations include extension specification languages for formal extension definition, compatibility testing frameworks for extension validation, security analysis tools for extension security assessment, and registry systems for extension publication and discovery.

Extension design should follow modular architecture principles, enabling independent development and deployment of protocol enhancements while maintaining core protocol security and functionality. Standardization efforts should establish common extension patterns, security requirements, and interoperability guidelines to promote ecosystem coherence.

\subsection{Scalability and Performance}
ACNBP implementation must address scalability challenges inherent in large-scale multi-agent systems. Performance optimization strategies include protocol message optimization for reduced bandwidth usage (e.g., using efficient data formats over verbose ones), caching systems for frequently accessed information, parallel processing for concurrent negotiations, and load distribution mechanisms for high-volume scenarios.

Implementation should support horizontal scaling through distributed protocol state management, stateless protocol design where possible, and efficient resource utilization patterns. Performance monitoring and optimization should be integrated throughout the implementation to enable continuous improvement and capacity planning.

\subsection{Cost and Payment Integration}
Many ACNBP implementations will require integration with payment and billing systems for commercial capability provision. Implementation considerations include secure payment protocol integration (e.g., using blockchain \cite{garcia2025blockchain}), micropayment systems for fine-grained capability pricing, escrow mechanisms for capability payment assurance, and billing integration for enterprise deployments.

Payment integration should support multiple payment models including pay-per-use, subscription-based, and auction-based pricing. The implementation should also provide cost prediction and budgeting capabilities to help requester agents manage capability acquisition costs within specified budgets.

\subsection{Error Handling and Resilience}
Robust error handling and resilience mechanisms are essential for reliable ACNBP operation in distributed environments. Implementation should include comprehensive error classification and handling procedures, timeout and retry mechanisms for network reliability, graceful degradation strategies for partial system failures, and recovery procedures for protocol state consistency.

Resilience mechanisms should address various failure scenarios including network partitions, agent failures, ANS unavailability, and resource exhaustion. The implementation should provide clear error reporting and diagnostic capabilities to support troubleshooting and system maintenance.

\subsection{Auditability and Logging}
Comprehensive audit trails and logging capabilities are essential for security, compliance, and system analysis. Implementation requirements include cryptographically protected audit logs, comprehensive event logging throughout protocol execution, privacy-preserving logging for sensitive operations, and integration with external audit and compliance systems.

Audit implementation should support forensic analysis, compliance reporting, and performance analysis while protecting sensitive information and maintaining system performance. Log aggregation and analysis capabilities should enable system-wide visibility and automated anomaly detection.

\section{Discussion}
ACNBP represents a significant advancement in agent communication protocols, addressing critical limitations of existing approaches while introducing novel mechanisms for secure, scalable, and verifiable agent interaction. This section analyzes the protocol's strengths, limitations, and position within the broader landscape of multi-agent system research.

\subsection{Protocol Strengths}
ACNBP's primary strengths include comprehensive security integration throughout the protocol design, scalable discovery mechanisms through ANS integration, flexible extension capabilities through protocolExtension, formal verification support through mathematical specification, and practical applicability demonstrated through detailed examples and implementation guidance.

The protocol's security-first design distinguishes it from previous approaches that treat security as an afterthought. By integrating cryptographic mechanisms, capability attestation, and comprehensive threat modeling throughout the protocol specification, ACNBP provides robust security foundations suitable for trust-critical applications.

The ANS integration provides scalable capability discovery that addresses a fundamental limitation of previous protocols. Unlike approaches that rely on broadcast or flooding mechanisms, ACNBP enables efficient, targeted capability discovery in large-scale agent populations through hierarchical, distributed directory services.

\subsection{Protocol Limitations}
Despite its comprehensive design, ACNBP has several limitations that may affect its applicability in certain scenarios. The protocol's complexity may create implementation challenges for simple agent systems that do not require sophisticated security or negotiation capabilities. The dependence on ANS infrastructure introduces potential single points of failure and requires significant infrastructure investment for deployment.

The protocol's security mechanisms, while comprehensive, introduce computational and communication overhead that may be excessive for resource-constrained environments or real-time applications with strict latency requirements. The capability verification and attestation mechanisms, while providing strong security guarantees, may be complex to implement and validate in practice.

\subsection{Comparison with Existing Protocols}
Compared to traditional protocols like CNP and FIPA, ACNBP provides significantly enhanced security, discovery, and extensibility capabilities while maintaining backward compatibility through protocolExtension mechanisms. Unlike semantic web service approaches, ACNBP is specifically designed for agent-to-agent interaction with support for autonomy, proactivity, and dynamic capability evolution.

Recent emerging protocols such as A2A \cite{A2A2025}, MCP \cite{MCPSpec2025}, and ACP \cite{zhang2025acp} address some limitations of traditional approaches but lack ACNBP's comprehensive integration of security, discovery, and negotiation mechanisms \cite{securing_a2a, narajala2025enterprise}. ACNBP's formal specification and security analysis provide stronger foundations for trust-critical applications than existing alternatives.

\subsection{Applicability and Adoption Considerations}
ACNBP is particularly well-suited for enterprise agent ecosystems, security-conscious applications, regulated industries requiring audit trails and compliance verification, and large-scale agent systems requiring efficient discovery and negotiation mechanisms. The protocol's complexity may limit adoption in simple agent systems or resource-constrained environments.

Successful adoption will likely require development of supporting tools, libraries, and infrastructure components to reduce implementation complexity and provide standard components for common use cases. Integration with existing agent frameworks and development environments will be critical for practical adoption.

\section{Future Work}
Several research directions emerge from the ACNBP specification and analysis, offering opportunities for protocol enhancement, theoretical development, and practical application extension.

\subsection{Formal Verification and Model Checking}
Future work should include comprehensive formal verification of ACNBP security properties using model checking tools and theorem proving systems. This includes verification of protocol correctness, security property validation, and automated analysis of protocol extensions for security and correctness preservation.

\subsection{Performance Optimization and Scalability Analysis}
Detailed performance analysis and optimization research is needed to characterize ACNBP behavior in large-scale deployments. This includes empirical scalability studies, performance optimization techniques, and comparative analysis with alternative protocols under various load conditions.

\subsection{Machine Learning Integration}
Integration of machine learning techniques for capability matching, agent selection optimization, and anomaly detection represents a promising research direction. This includes semantic capability matching using natural language processing, reinforcement learning for negotiation strategy optimization, and predictive models for capability performance and reliability.

\subsection{Blockchain and Distributed Ledger Integration}
Research into blockchain integration \cite{garcia2025blockchain} for capability verification, payment processing, and audit trail immutability could enhance ACNBP's trust and transparency characteristics. This includes smart contract integration for automated capability binding, cryptocurrency payment integration, and distributed reputation systems.

\subsection{IoT and Edge Computing Applications}
Adaptation of ACNBP for Internet of Things (IoT) and edge computing environments \cite{lee2024iot} presents interesting challenges including resource-constrained implementation, low-latency requirements, and intermittent connectivity scenarios. Research should address protocol simplification, offline operation modes, and energy-efficient implementation strategies.

\subsection{Regulatory Compliance and Governance}
Research into regulatory compliance frameworks \cite{miller2025compliance}, governance mechanisms, and policy integration for ACNBP deployment in regulated industries is needed. This includes compliance automation, regulatory reporting integration, and governance frameworks for agent ecosystem management.

\subsection{Cross-Protocol Interoperability}
Development of interoperability mechanisms between ACNBP and existing agent communication protocols could facilitate gradual adoption and ecosystem integration. This includes protocol translation mechanisms, gateway systems, and migration strategies for existing agent systems.

\subsection{Advanced Threat Modeling and Security Analysis}
Continued security research should address emerging threats, advanced persistent threats targeting agent systems, and security implications of protocol extensions. This includes development of agent-specific security metrics, threat intelligence integration, and automated security analysis tools based on frameworks like MAESTRO \cite{mas_threat_model_2025}.

\section{Conclusion}
The Agent Capability Negotiation and Binding Protocol (ACNBP) presents a comprehensive solution to critical challenges in heterogeneous multi-agent systems through its integration of secure capability discovery, sophisticated negotiation mechanisms, and verifiable binding commitments. The protocol's security-first design, scalable ANS integration, and extensible architecture position it as a significant advancement over existing agent communication protocols.

ACNBP's formal specification provides strong theoretical foundations while its practical implementation guidance and detailed examples demonstrate real-world applicability. The comprehensive security analysis using the MAESTRO framework addresses the full spectrum of threats relevant to agent communication, providing confidence for deployment in trust-critical applications.

The protocol's protocolExtension mechanism ensures backward compatibility and supports ecosystem evolution, addressing a critical limitation of previous approaches that often required wholesale replacement for enhancement or adaptation. This extensibility, combined with the protocol's formal specification, enables gradual adoption and continuous improvement based on operational experience and evolving requirements.

While ACNBP introduces complexity compared to simpler protocols, this complexity is justified by the sophisticated requirements of modern heterogeneous agent ecosystems. The protocol's comprehensive approach to security, discovery, negotiation, and binding provides essential capabilities for enterprise deployment, regulated applications, and large-scale agent systems.

Future research directions offer opportunities for continued protocol enhancement, theoretical development, and practical application extension. The protocol's formal foundations provide a solid basis for verification, optimization, and extension research that can further enhance its effectiveness and broaden its applicability.

ACNBP represents a significant step forward in enabling secure, efficient, and verifiable collaboration between autonomous agents in diverse, dynamic, and potentially adversarial environments. Its comprehensive approach to the challenges of agent capability negotiation positions it as a foundational protocol for the next generation of multi-agent systems.

%
% --- Optional Sections ---
%
\section*{Intellectual Property and Copyright Compliance}
This AI research paper and all associated work comply fully with intellectual property laws and copyright regulations. The authors declare that:
\begin{itemize}
\item All original research, concepts, and methodologies presented in this work are the intellectual property of the authors and have been developed independently.
\item All referenced works, prior art, and external sources have been properly cited and attributed in accordance with academic standards and copyright requirements.
\item The Agent Capability Negotiation and Binding Protocol (ACNBP) represents novel contributions to the field of multi-agent systems and does not infringe upon existing patents or proprietary technologies.
\item All software concepts, algorithms, and technical specifications described herein are presented for academic and research purposes under fair use provisions.
\item The authors acknowledge and respect the intellectual property rights of all cited works and have made every effort to ensure accurate attribution and compliance with copyright laws.
\item This work has been conducted in accordance with the intellectual property policies of the authors' respective institutions: DistributedApps.ai, Cisco Systems, Amazon Web Services, and Intuit.
\item The research methodology, experimental design, and analysis presented are original contributions that advance the state of knowledge in agent communication protocols.
\item Any potential commercial applications or implementations of ACNBP should be developed in consultation with the authors and in compliance with applicable intellectual property laws.
\end{itemize}
The authors affirm their commitment to maintaining the highest standards of academic integrity and intellectual property compliance in all aspects of this research.

\section*{Acknowledgment}
The authors would like to thank the anonymous reviewers for their insightful feedback and valuable suggestions, which greatly improved the quality of this paper. This work was supported in part by the authors' respective institutions.

%
% --- BIBLIOGRAPHY ---
%
\bibliographystyle{IEEEtran}
\bibliography{references}

\end{document}